\PassOptionsToPackage{table}{xcolor}
\documentclass[sigconf]{acmart}
\usepackage{etoolbox}
\undef\Bbbk

\usepackage{amsfonts}
\usepackage{amssymb}

\usepackage{bbm}
\usepackage{bm}

\usepackage{algorithm}
\usepackage{graphicx}
\usepackage{subcaption}
\usepackage{multirow}
\usepackage{hyperref}
\usepackage{algpseudocode}

\AtBeginDocument{%
  }

\copyrightyear{2026}
\acmYear{2026}
\setcopyright{cc}
\setcctype{by}
\acmConference[WWW '26]{Proceedings of the ACM Web Conference 2026}{April 13--17, 2026}{Dubai, United Arab Emirates}
\acmBooktitle{Proceedings of the ACM Web Conference 2026 (WWW '26), April 13--17, 2026, Dubai, United Arab Emirates}
\acmPrice{}
\acmDOI{10.1145/3774904.3792202}
\acmISBN{979-8-4007-2307-0/2026/04}

\begin{document}

% \title{LBM: Offline Reinforcement Fine-Tuning a Large Language Model for Generative Auto-Bidding}
% \title{Mastering Auto-Bidding via A New Generative Backbone based on Large Language Model}
\title{LBM: Hierarchical Large Auto-Bidding Model \\via Reasoning and Acting}
% \author{Yewen Li}
% \email{yewen001@e.ntu.edu.sg}
% \author{Shuai Mao}
% \affiliation{%

% \author{
%   Author One\thanks{These authors contributed equally to this work.} \\
%   Affiliation 1 \\
%   \texttt{email1@example.com}
%   \and
%   Author Two\footnotemark[1] \\
%   Affiliation 2 \\
%   \texttt{email2@example.com}
% }

\author{Yewen Li}
\authornote{Both authors contributed equally to this research.}
\affiliation{%
  \institution{Kuaishou Technology}
  \city{Beijing}
  % \state{Beijing Shi}
  \country{China}
  }
\email{liyewen@kuaishou.com}

\author{Zhiyi Lyu}
\authornotemark[1]
\affiliation{%
  \institution{Nanyang Technological University}
  \city{Singapore}
  % \state{Beijing Shi}
  \country{Singapore}
  }
\email{zhiyi.lyu@ntu.edu.sg}

\author{Peng Jiang}
% \authornote{Corresponding author.}
\affiliation{%
  \institution{Kuaishou Technology}
  \city{Beijing}
  \country{China}
}
\email{jiangpeng07@kuaishou.com}

\author{Qingpeng Cai}
\authornote{Corresponding author.}
\affiliation{%
  \institution{Kuaishou Technology}
  \city{Beijing}
  \country{China}
}
\email{caiqingpeng@kuaishou.com}

\author{Fei Pan}
% \authornote{Corresponding author.}
\affiliation{%
  \institution{Kuaishou Technology}
  \city{Beijing}
  \country{China}
}
\email{panfei05@kuaishou.com}

\author{Bo An}
\affiliation{%
  \institution{Nanyang Technological University}
  \city{Singapore}
  \country{Singapore}
  }
\email{boan@ntu.edu.sg}

\author{Peng Jiang}
% \authornote{Corresponding author.}
\affiliation{%
  \institution{Kuaishou Technology}
  \city{Beijing}
  \country{China}
}
\email{jiangpeng@kuaishou.com}

% \renewcommand{\shortauthors}{Li et al.}
\renewcommand{\shortauthors}{Yewen Li et al.}
\begin{abstract}
% 生成式自动出价越来越重要 --> 现有的生成式出价方法只在DT和Diffuser上做，然而出价是一个有着强逻辑的任务，专家知识是可以很大程度影响决策性能，提升泛化能力，然而现有方法并未探究LLM的推理能力对出价的帮助 --> 我们发现存在问题：1）纯文本输入让LLM出价 性能很差且训练低效，说明LLM无法做精确控制任务；2）LLM并未在出价任务上预训练，其CoT的质量比较低，然而GRPO等提升cot能力的方法无法进行online探索rollout得到reward信号，无法直接应用于offline任务 
% 我们提出方法 GAB-Delta_Q: 引入decision_embedding的概念融合文本的逻辑能力和数值序列的精准能力，性能能比DT/Diffusion等backbone更好; 基于offline RL引入Delta-Q衡量CoT对决策的影响，并基于此做GRPO做RFT提升CoT对决策的增益能力；
% 我们在离线数据集上做了验证，证明LLM可以作为一种新的生成式backbone提升Auto-bidding的能力

The growing scale of ad auctions on online advertising platforms has intensified competition, making manual bidding impractical and necessitating auto-bidding to help advertisers achieve their economic goals. Current auto-bidding methods have evolved to use offline reinforcement learning or generative methods to optimize bidding strategies, but they can sometimes behave counterintuitively due to the black-box training manner and limited mode coverage of datasets, leading to challenges in understanding task status and generalization in dynamic ad environments. 
Large language models (LLMs) offer a promising solution by leveraging prior human knowledge and reasoning abilities to improve auto-bidding performance. However, directly applying LLMs to auto-bidding faces difficulties due to the need for precise actions in competitive auctions and the lack of specialized auto-bidding knowledge, which can lead to hallucinations and suboptimal decisions. 
To address these challenges, we propose a hierarchical \textbf{L}arge auto-\textbf{B}idding \textbf{M}odel (\textbf{LBM}) to leverage the reasoning capabilities of LLMs for developing a superior auto-bidding strategy. This includes a high-level LBM-Think model for reasoning and a low-level LBM-Act model for action generation. Specifically, we propose a dual embedding mechanism to efficiently fuse two modalities, including language and numerical inputs, for language-guided training of the LBM-Act; then, we propose an offline reinforcement fine-tuning technique termed GQPO for mitigating the LLM-Think's hallucinations and enhancing decision-making performance without simulation or real-world rollout like previous multi-turn LLM-based methods. 
Experiments demonstrate the superiority of a generative backbone based on our LBM, especially in an efficient training manner and generalization ability.\footnote{Code is available at https://github.com/yewen99/LBM-WWW26.}
\end{abstract}

%    <concept>
%        <concept_id>10010147.10010178.10010199.10010201</concept_id>
%        <concept_desc>Computing methodologies~Planning under uncertainty</concept_desc>
%        <concept_significance>500</concept_significance>
%        </concept>
%  </ccs2012>

% \ccsdesc[500]{Computing methodologies~Planning under uncertainty}





\maketitle

\section{Introduction}

\begin{figure*}[t!]
    \centering
    \includegraphics[width=\linewidth]{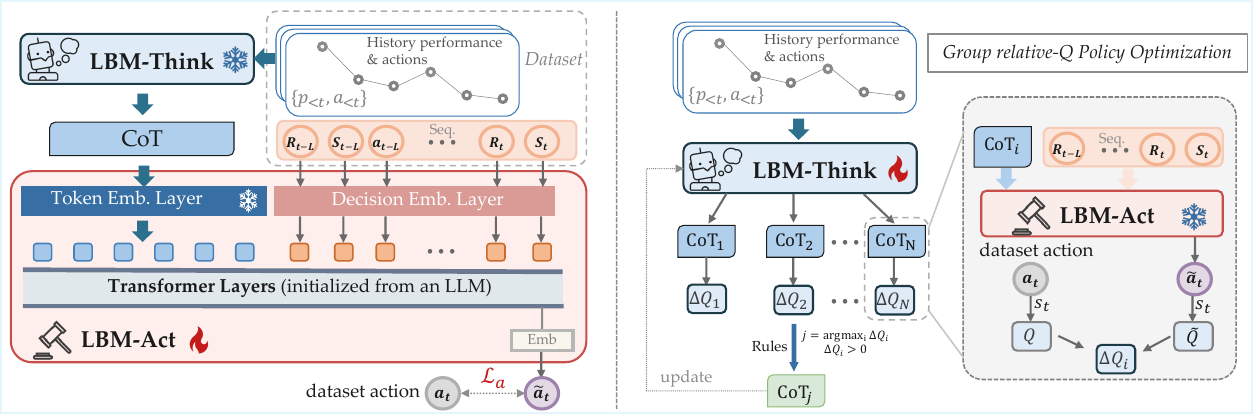}
    \caption{Overview of the training scheme for our hierarchical large auto-bidding model. \textbf{Training Stage I} (left): train the LBM-Act with language-guided decision training via a dual embedding mechanism to fuse two modalities; \textbf{Training Stage II} (right): reinforcement fine-tuning the LBM-Think via group relative-Q policy optimization.}
    \label{fig:main}
    \vspace{-3mm}
\end{figure*}

% {\yw[Para 1: What is auto-bidding and introduce RL and Generative methods.]}
The rapid digitization of commerce has significantly expanded the reach of online advertising platforms on the web, leading to a surge in advertisers competing for millions or even billions of impression opportunities through ad auctions \cite{DBLP:conf/wsdm/WangY15, evans2009online}. 
Previously, advertisers relied on experienced human experts to manually adjust bids, but it is becoming impractical due to the increasing competitiveness of auctions \cite{emerge}.
Therefore, auto-bidding has become essential for helping advertisers achieve their commercial goals in the face of competition \cite{BM}.
Auto-bidding can be modeled as a sequential decision-making task, where the objective is to maximize conversions while adhering to specific economic constraints, such as cost-per-action (CPA), by adjusting bidding parameters over a bidding period \cite{sorl, USCB}.
% As we can easily collect bidding logs to create an offline training dataset from online advertising platforms, current auto-bidding methods are built on offline reinforcement learning (RL) methods or generative methods. These methods aim to learn an optimal policy for achieving reward-maximizing trajectories by stitching together sub-optimal trajectories from datasets.
Current auto-bidding methods are mainly built on offline reinforcement learning (RL) or generative methods, leveraging bidding logs collected from online advertising platforms \cite{gas,gave}. 
These methods aim to learn optimal policies by stitching sub-optimal trajectories from datasets to achieve reward-maximizing trajectories \cite{diffuser}.
Offline RL typically performs trajectory stitching using dynamic programming with a value function \cite{iql}. 
Generative auto-bidding methods, which do not rely on a value function, perform stitching through a conditional generation approach utilizing the powerful generative models \cite{dt, diffuser, diffuser_0}. For instance, the Decision Transformer (DT) employs an auto-regressive transformer to generate bid actions based on historical sequences that encompass return-to-go, states, and actions \cite{dt, cdt}.

However, these RL and generative methods can occasionally act counterintuitively, such as increasing bidding parameters even when the CPA significantly exceeds the constraint, which is unlikely to occur in manual bidding.
This issue arises because these methods primarily rely on reward design to acquire desired properties \cite{diffuser, diffuser_0, cbd}. Consequently, it becomes challenging to ensure that the model fully understands the task status, resulting in a fully black-box approach. 
Additionally, since it's impossible to capture every corner case into datasets, these methods are inherently restricted by the mode coverage of the offline dataset and lack guarantees for generalization performance in dynamic advertising environments, especially in unforeseen situations \cite{gas}.
These limitations present significant challenges and could be major obstacles to the progress and adoption of advanced auto-bidding services, as frequent abnormal model behavior can undermine the trust of advertisers \cite{display_ads}. 
Fortunately, large language models (LLMs) are revolutionizing decision-making by harnessing prior human knowledge to boost performance \cite{ qwen, gpt-2}. 
These LLMs effectively follow human instructions and utilize their reasoning ability with pre-trained knowledge to comprehend the task status and reasoning for optimal actions, showcasing superior performance and generalization ability in tasks like math and code \cite{guo2025deepseek}. 
Thus, incorporating LLMs as a new generative backbone for auto-bidding presents promising opportunities for improvement.

% Thus, incorporating LLMs as a new generative backbone for auto-bidding presents promising opportunities for improvement.

% {\yw[Para 3: What is the difficulty in directly applying LLM to auto-bidding?]
% }
While LLMs can be directly applied to auto-bidding by converting sequential numerical information into language token inputs and generating bidding parameters as a response, their performance is constrained via prompt engineering \cite{in_context, cot} in large-scale auctions due to the generated suboptimal actions. 
This limitation arises from the intense competition among advertisers, where their bids can be very close, and thus, the need for precise actions is critical for optimality \cite{aigb,auctionNet}. 
A suboptimal bidding parameter could lead to budget waste or limited gains in impressions. For instance, if the bidding parameter is set too high, the budget may be exhausted fast, missing later opportunities; conversely, if set too low, it may result in no acquired impressions.
To our knowledge, current public LLMs have not been specially pretrained or fine-tuned on auto-bidding data, as there is no public auto-bidding data in text format; thus, they could still suffer from the notorious hallucinations issue and lack optimal sequential decision-making ability \cite{gigpo, gspo}.

% {\yw[Para 4: How previous works address the issues of LLM-based methods?]}
There have been many attempts to enhance LLMs for decision-making tasks, though they are not well-suited for the auto-bidding task. 
To leverage the reasoning capabilities of LLMs, some approaches, such as reinforcement learning with verifiable rewards (RLVR) \cite{guo2025deepseek}, aim to enhance performance in tasks like mathematical reasoning, code generation, and computer control \cite{gigpo}. These methods primarily focus on reasoning abilities rather than continuous control, where the action space is typically discrete, such as in tool-use agents \cite{digi-q}, whereas auto-bidding operates in a continuous numerical action space.
For applications of LLMs in tasks with continuous state and action spaces, there is one challenge that representing numerical inputs as language tokens can be costly \cite{llm-dt, llm-dt-1}; for instance, a numerical state like "12.34" requires five tokens, potentially resulting in thousands of tokens when using extensive historical data in complex tasks. Even if it is done in this way, the attention mechanism of LLMs could be limited, leading to hallucinations and suboptimal performance \cite{long-context, long-context-1}.
For addressing this, LLM-DT \cite{llm-dt} fine-tunes the LLM with additional embedding layers that process long numerical sequences into more informative representations.
However, this approach actually treats the LLM as a Decision Transformer (DT) initialized with transformer layers from the LLM, which limits the full exploitation of the LLM's knowledge and reasoning capabilities.

% Therefore, it is 
 
% \begin{center}
% \fbox{\begin{minipage}{0.9\linewidth}
% \textit{Can we utilize the knowledge and reasoning ability of LLM for achieving an optimal and explainable bidding strategy in an offline training manner?}
% \end{minipage}}
% \end{center}

% {\yw[Para 5: How we address the issues and Summarize of our contribution.]}
To leverage the reasoning capabilities of LLMs for developing an optimal auto-bidding strategy, we introduce a hierarchical \textbf{L}arge auto-\textbf{B}idding \textbf{M}odel (\textbf{LBM}), which comprises two modules: i) the LBM-Think module, which handles high-level reasoning in the language space, and ii) the LBM-Act module, which focuses on low-level decision-making in the continuous action space.
Specifically, the LBM-Think module is tasked with generating a chain-of-thought (CoT) \cite{cot} in language format that summarizes and reflects historical bidding status, while reasoning about high-level future actions like adjustment directions.
Note that CoT can be pre-generated asynchronously ahead of the next decision timestep as shown in Fig. \ref{fig:inference}, making it suitable for industry applications.
For the LBM-Act module, we introduce a \textbf{{dual embedding}} mechanism to encode both the numerical sequence of observations and language CoT. 
% Transformer layers from a smaller pretrained LLM, then project these fused embeddings into a final representation. The final bidding action is determined by processing this output representation through an MLP.
This hierarchical structure enables the LBM to comprehend task status while executing precise control efficiently.
Additionally, it separates CoT generation and action execution into distinct LLM backbones, with LBM-Think utilizing a larger LLM and LBM-Act employing a smaller LLM. 
For training such a hierarchical LBM, we propose a two-stage training scheme, as illustrated in Fig.~\ref{fig:main}.
In the first stage, we introduce a language-guided decision training approach with additional embedding layers to train the LBM-Act model, allowing it to effectively fuse two modalities of inputs and generate precise actions. 
In the second stage, we propose a Group relative-Q Policy Optimization (\textbf{{GQPO}}) method to fine-tune the LBM-Think model, reducing hallucinations and achieving better reasoning performance in a stable offline manner. 
% Specifically, LBM-Think generates a group of CoTs, and the trained LBM-Act model uses these CoTs to generate actions. A trained Q-value is then used to assess whether a CoT leads to an action with incremental effectiveness compared to the dataset action, represented as relative-Q $\Delta Q$. 
This relative-Q serves as an assessment of the effectiveness of a CoT to fine-tune the LBM-Think entirely offline, unlike previous multi-turn LLM-based agents that require rollout in simulators or the real world, which is not feasible and can be risky for auto-bidding \cite{gigpo, digi-q}.
To summarize, the contributions of this work are:
\begin{itemize}
    \item To leverage the reasoning capabilities of LLMs for developing an advanced auto-bidding strategy, we propose a novel hierarchical \textbf{L}arge auto-\textbf{B}idding \textbf{M}odel (\textbf{LBM}) based on LLMs, including a high-level LBM-Think for reasoning and a low-level LBM-Act for precise action generation;
    \item We propose a \textbf{dual embedding} mechanism to efficiently fuse two modalities, including language and numerical inputs, for language-guided training of the LBM-Act;
    \item We propose a stable fine-tuning technique termed \textbf{GQPO} for mitigating the LLM-Think's hallucinations and enhancing decision-making performance in an offline manner;
    \item Experiments demonstrate the superiority of a generative backbone based on LLMs, especially regarding the efficient training manner and generalization ability.
\end{itemize}

\vspace{-2mm}
\section{Preliminary}
\subsection{Problem Statement}
Consider a scenario where impression opportunities arrive sequentially, each identified by an index \(i\). An advertiser secures an impression if its bid \(b_i\) exceeds those of other advertisers, resulting in a cost \(c_i\). The objective is to maximize the cumulative value of the impressions won, expressed as \(\sum_i o_i v_i\), where \(v_i\) represents the value of the impression and \(o_i\) is a binary variable indicating whether the advertiser wins impression \(i\). Furthermore, it is crucial to account for budgetary and various economic constraints, such as limiting the unit cost for specific advertising events like CPC and CPA \citep{USCB}. For simplicity, we focus on auto-bidding with cost-related constraints, which can be uniformly formulated as:
$
\frac{\sum_i c_{ij} o_i}{\sum_i p_{ij} o_i} \leq C_j,
$
where \(C_j\) is the upper bound of \(j\)-th constraint provided by the advertiser. \(p_{ij}\) can be any performance indicator, \textit{e.g.}, return or constant. \(c_{ij}\) is the cost of constraint \(j\). Therefore, the objective of auto-bidding is:
\begin{equation}
\begin{aligned}
\text{maximize} \quad & \sum_i o_i v_i \\
\text{s.t.} \quad & \sum_i o_i c_i \leq B  \\
& \frac{\sum_i c_{ij} o_i}{\sum_i p_{ij} o_i} \leq C_j, \quad \forall j  \\
& o_i \in \{0, 1\}, \quad \forall i
\label{eq:goal}
\end{aligned}
\end{equation}
A previous study \cite{USCB} has already shown the optimal solution:
\begin{equation}
b_i^* = \lambda_0 v_i + \sum_{j=1}^{J} \lambda_j p_{ij}{C_j},
\label{eq:optimal_bid}
\end{equation}
where \(b_i^*\) is the optimal bid for impression \(i\). 
The parameters \(\lambda_j\) represent the optimal {bidding parameters}. 
Please note that in industrial ad platforms, \textbf{current auto-bidding methods mainly focus on adjusting bidding parameters at a low frequency, such as every 30 minutes, rather than for every single impression that arises in millisecond} \cite{auctionNet}.
Therefore, there is ample time for an advanced auto-bidding method to respond to bidding parameter adjustment requirements.

\subsection{Decision-making Process for Auto-bidding}
Given the dynamic nature of the advertising environment, optimal bidding parameters must be regularly adjusted to maximize total value, framing the auto-bidding task as a sequential decision-making process.
% We consider a standard decision-making setup consisting of an auto-bidding agent interacting with advertising environment ${E}$ in discrete timesteps.
At each timestep $t$ of a bidding period, the agent receives a state $s_t\in\mathcal{S}$ describing the real-time advertising status and then outputs an action $a_t\in\mathcal{A}$ for the final bid.
% {\yw Note that, this is more flexible and general than the previous definition of the action, \textit{i.e.}, adjustment to the bidding parameters at the time period \(t\), which is restricted to the specific dimension of the number of bidding parameters \(\lambda_j\), \(j = 0, \ldots, J\) and modeled as \((a_t^{\lambda_0}, \ldots, a_t^{\lambda_J})\).}
The advertising environment has an underlying unknown state transition dynamic $\mathcal{T}$.
% Under an MDP assumption, the transition dynamic $\mathcal{T}$ could be expressed by $\mathcal{T}:s_t \times a_t \rightarrow s_{t+1}$, \textit{i.e.}, the next state $s_{t+1} \in \mathcal{S}$ is determined by the current state $s_t$ and action $a_t$.
% In this case, the agent's policy is expressed as $\pi(a_t|s_t)$.
% Without an MDP assumption, 
The next state could be determined by both the historical information and current observations.
After transitioning to the next state, the advertising environment would emit a reward $r_t $ representing the value contributed to the objective obtained within the time period $t$.
This process would repeat until the end of a bidding period, such as one day.
% Repeating this process until the end of a bidding period, one day for instance, the goal of the auto-bidding agent is to maximize the total received value over the entire period as stated in Equation \ref{eq:goal}.
A detailed description of our modeling is listed below:
\begin{itemize}
    \item \textbf{$s_t$}: the state is a collection of information that describes the advertising status from the perspective of a campaign. The information should in principle reflect time, budget consumption and KPI constraints satisfying status, such as left time, left budget, budget consumption speed, current KPI ratio for constraint \(j\) (i.e. \((\Sigma_i c_{ij} o_i / \Sigma_i p_{ij} o_i) / C_j\)), etc.
    \item \(a_t\): adjustment to the bidding parameters \(\lambda_j\), \(j = 0, \ldots, J\), at the time period \(t\),
    and modeled as \((a_t^{\lambda_0}, \ldots, a_t^{\lambda_J})\). 
    \item $r_t$ : a typical setting of the reward could be the conversion value contributed to the objective obtained within the time period from $t$ to $t+1$.
    % \(r_t = \Sigma_{i \in \mathcal{C}} o_i v_i\), i.e. the sum value of the winning impressions in \(\mathcal{C}\). 
    % For simplicity, we will omit $\mathcal{C}$ in the following sections.
\end{itemize}

\vspace{-3mm}
\section{LBM: Hierarchical Large Auto-Bidding Model}
In this section, we first explain how we construct the hierarchical Large auto-Bidding Model (LBM) with two modules: high-level LBM-Think and low-level LBM-Act. Next, we propose a two-stage training method for the model. Initially, we introduce a language-guided decision training method for LBM-Act. To further enhance the LBM's performance, particularly the reasoning ability of the LBM-Think, we then present the GQPO method, which utilizes an offline Q-value for fine-tuning it.

\vspace{-3mm}
\subsection{Model Structure and Inference}
% {\yw{[Motivation for adopting a hierarchical architecture]}}
To improve reasoning capabilities, previous methods expand models to incorporate a large number of parameters, often surpassing 1 billion. This approach can be costly and result in slow inference for industrial applications, especially if each decision timestep requires immediate reasoning and action using LLMs.
Even if we aim to create a model capable of both reasoning and generating optimal actions in a continuous action space, training such a model is challenging and may degrade the original model's general reasoning ability. This is illustrated in { Fig. \ref{fig:train_response_length}}, where we use the GRPO to fine-tune an LLM for action generation with CoT.
Therefore, a practical and flexible approach is to propose a hierarchical structure that decouples reasoning and acting into two separate modules. Reasoning capabilities can leverage existing LLMs to process historical information asynchronously, without needing a response at every decision timestep. The acting capability can be built on a smaller LLM that focuses on immediate decision-making using the pre-generated reasoning output and current state information.

% {\yw [Can LBM-Act do both reasoning and action --> GRPO failure]}

% {\yw[Details of the inference process]}
% We now give a detailed inference procedure of our hierarchical large auto-bidding model.
For sequential inference procedure of auto-bidding, we can observe a sequence of historical information with horizon length $H$, \textit{i.e.}, $\{p_{i}, a_i\}_{i=t-H:t-1}$, where $p_i$ is the key performance indicators reflecting the bidding status like the achieved conversions, budget utilization ratio, and CPA. For every decision-making step, we have a state $s_t$ representing the current detailed observation, like the impression value distribution and auction status.
For the high-level LBM-Think model, denoted as $h_\theta$, it summarizes and reflects the historical sequential information only and then reasons for a future high-level decision like the adjustment direction of the bidding parameters, producing a CoT denoted as $c_t$, \textit{i.e.},
\begin{align}
   \text{\textbf{LBM-Think}:}\quad c_t =h_\theta(\{p_{i}, a_i\}_{i=t-H:t-1}).
   \label{eq:lbm-think}
\end{align}
Note that this high-level process could be done asynchronously before timestep $t$, and we could simplify the performance indicator $p$ with only the necessary elements from states, thus leaving sufficient time for the LLM to generate. 
After receiving CoT $c_t$ and current state $s_t$, the LBM-Act model $f_\phi$ generates an action $\tilde{a}_t$, \textit{i.e.},
\begin{align}
    \text{\textbf{LBM-Act}:}\quad \tilde{a}_t = f_\phi(s_t, c_t). 
    \label{eq:lbm-act}
\end{align}
To enhance the decision-making performance, we can extend the $s_t$ to include more information in a DT manner, including \textit{return-to-go} $R_{\leq t}$, states $s_{\leq t}$, and actions $a_{< t}$.
Therefore, the LBM-Act needs to fuse two modalities of inputs, including language and number.
% As the $c_t$ are language modality and $s_t$
A detailed illustration of LBM could be seen in Fig. \ref{fig:main} and Fig. \ref{fig:inference}.

% {\yw [derive to two-stage training]}
Given that different modules have distinct objectives, where LBM-Think $h_\theta$ focuses on understanding the bidding status and LBM-Act $f_\phi$ concentrates on low-level action generation, we propose a stable two-stage training approach. 
Specifically, this involves first training the LBM-Act through language-guided decision training using a dual embedding mechanism, followed by reinforcement fine-tuning of the LBM-Think with the trained LBM-Act.

\subsection{Language-guided Decision Training for LBM-Act via Dual Embedding}

% {\yw [Combine two modalities via a Dual Embedding]}
While it's possible to convert numerical states into language format, this approach is inefficient and can consume a significant number of tokens. For example, representing "12.34" in language format can require 5 tokens, and thus handling sequences of high-dimensional states could result in the use of thousands of tokens. Additionally, the performance of attention mechanisms is often constrained when processing long sequences of token inputs, particularly when using smaller language models for LBM-Act, which can lead to suboptimal performance in auto-bidding scenarios \cite{long-context, long-context-1}.
To address these challenges, we propose a \textbf{dual embedding} mechanism to fuse these two modalities, rather than unifying them solely in the language modality. Dual embedding involves employing two distinct embedding layers to process data: one for language modality using a pretrained token embedding layer for the CoT generated by the LBM-Think model, and another for numerical data, like observed numerical sequences, using a \textit{decision embedding layer}. Each numerical state is projected into a single embedding using an additional MLP, termed "decision embedding", which matches the size of a token embedding and encapsulates decision-related information.
By integrating these two embeddings, LBM-Act then employs transformer layers initialized with the pretrained weight of an LLM to project them to the final layer's output representation, and a final action is determined via an MLP processing this representation.
Therefore, with this dual embedding mechanism, the LBM-Act can effectively comprehend language instructions and manage complex decision-making tasks.

% {\yw [A training scheme]}
To enhance the LBM-Act model's ability to handle two modalities, we propose a language-guided decision training method. Specifically, we first utilize the LBM-Think module to generate a CoT, which involves summarizing historical bidding performance information and reasoning for optimal future high-level decisions, such as determining the direction for bidding parameter adjustments. Additionally, we obtain a corresponding numerical sequence $\{R_{t-L:t}, s_{t-L:t}, a_{t-L:t-1}\}$ of length $L$, representing more comprehensive and detailed information.
In previous DT methods, the model learns to generate $a_t$ based on this sequence, aiming to achieve the current return-to-go $R_t$ during the training stage, rather than learning to generate an optimal action like offline RL methods such as IQL. 
However, there is uncertainty in achieving the goal represented by $R_t$, expressed as $p(a_t) = f(R_{\leq t}, s_{\leq t}, a_{<t}) = a_{t-1} + \epsilon$, where $\epsilon \in \mathbb{R}$ represents the uncertainty. It is not always clear which action aligns with the direction of the optimal action. 
For example, a favorable $R_t$ could be achieved with a bad $a_t$ but good subsequent actions $a_{>t}$.
To address potential conflicts between the two modalities during training, we use the direction of the dataset action $a_t$ compared to $a_{t-1}$ as an anchor direction. 
If a reasoned high-level direction in the CoT conflicts with this anchor direction, it will be disregarded. 
The accepted CoT component is concatenated with the numerical sequence $\{R_{t-L}, s_{t-L}, a_{t-L}, \ldots, R_t, s_t\}$ and sent to the dual embedding to obtain their corresponding embeddings $\bm{z} = \{\bm{z}_{cot}, \bm{z}_{s}\}$. The inner transformer layer learns to fuse these two modalities to generate actions using the attention mechanism, \textit{i.e.}, 
$
    \text{Attention}(\bm{Q}, \bm{K}, \bm{V}) = \text{softmax}\left(\frac{\bm{Q}\bm{K}^T}{\sqrt{d_k}}\right)\bm{V},
$
where $\bm{Q} = \bm{z}\bm{W}^Q$, $\bm{K} = \bm{z}\bm{W}^K$, and $\bm{V} = \bm{z}\bm{W}^V$. The output prediction action $\tilde{a}_t$, based on the fused representation, is trained to match the labeled action $a_t$ in the dataset, \textit{i.e.},
\begin{align}
    \mathcal{L}_a(\phi) = \left\|\tilde{a}_t - a_t\right\|_2.
\end{align}
Upon convergence, the LBM-Act can comprehend language guidance, follow its instructions, and achieve Return-to-go in continuous action space. During the inference stage, the return-to-go is typically initialized to its maximum value at the beginning timestep, aiming to achieve an optimal bidding strategy that aligns with the objective of LBM-Think.

% {\yw High-level }

% To improve the auto-bidding performance of LLM-based methods, especially in handling the precision action requirement in competitive auctions, we introduce a new structure for incorporating the high-level semantic ability of language and the low-level precise decision ability of numerical state information, which is novelly achieved via introducing a concept of ``decision embedding''.

\vspace{-2mm}
\subsection{Offline Reinforcement Fine-Tuning for LBM-Think via GQPO}
% {\yw [review previous RFT method --> need offline]}
Since existing LLMs have not been extensively pretrained on auto-bidding datasets in language format, they may exhibit hallucinations and demonstrate a suboptimal bidding strategy.
Previous related works on decision-making using LLMs have widely employed reinforcement fine-tuning techniques for enhanced performance \cite{guo2025deepseek,gspo,gigpo,digi-q}. 
A representative work is the group relative policy optimization (GRPO) \cite{guo2025deepseek}, expressed as 
{\small
\begin{align}
&\mathcal{J}_{\text{GRPO}}(\theta) = \mathbb{E}_{q \sim P(Q), \{o_i\}_{i=1}^{G} \sim \pi_{\theta_{\text{old}}}(O|q)} \\
&\frac{1}{G} \sum_{i=1}^{G} \frac{1}{|o_i|} \sum_{t=1}^{|o_i|} \left\{ \min \left[ \rho_{i,t} \hat{A}_{i,t}, \text{clip} \left( \rho_{i,t}, 1\pm\varepsilon \right) \hat{A}_{i,t} \right] - \beta \mathcal{D}_{\text{KL}} \left[ \pi_{\theta} \| \pi_{\text{ref}} \right] \right\}, \notag
\end{align}
}
where $\rho_{i,t}= \frac{\pi_{\theta}(o_{i,t}|q,o_{i,<t})}{\pi_{\theta_{\text{old}}}(o_{i,t}|q,o_{i,<t})}$ is the importance sampling weight, $\pi_\theta$ and $\pi_{\theta_{\text{old}}}$ represent the current and previous policy models, respectively. The variables $q$ and $o$ denote questions and outputs sampled from the question dataset and the old policy $\pi_{\theta_{\text{old}}}$. The parameter $\epsilon$ is a clipping-related hyperparameter to stabilize training. 
The advantage $\hat{A}_t$ is the advantage calculated based on the relative rewards of the outputs inside each group.
It needs a verifiable outcome-based reward function to score the outputs $\{o_i\}_{i=1}^{G}$, yielding $G$ rewards $r = \{r_1, r_2,...,r_G\}$ correspondingly.
GRPO could be directly extended to multi-turn decision-making tasks, leading to methods like GiGPO \cite{gigpo}.
However, these methods rely on rollouts in a simulator or the real world to obtain correct rewards, which is not feasible and could be risky in the context of auto-bidding. 
Consequently, it is necessary to explore a reinforcement fine-tuning approach that operates entirely offline. 

The primary challenge lies in evaluating the impact of CoT reasoning on decision-making performance without real-world rollouts.
Recall previous offline RL methods, the training objective is typically based on an advantage-weighted regression (AWR) technique for policy extraction \cite{awr_0, awr_1, awr_2, awr_3}, \textit{i.e.}, 
\begin{align}
    \mathcal{J}_{\text{AWR}} (\theta)=\mathbb{E}_{s,a\sim\mathcal{D}}[\exp(\beta(Q_\varphi(s,a)-V_\psi(s))\log \pi_\theta(a|s))],
    \label{eq:awr}
\end{align}
where $Q$ represents the state-action pair value, $V$ represents the state value that serves as a baseline for variance reduction, and $\beta\in[0,+\infty)$ is a hyperparameter.
In this work, we can train the value functions using a stable offline RL method named Implicit Q-Learning (IQL) \cite{iql}, formulated as:
\begin{align}
    \mathcal{L}_{V}(\psi) = \mathbb{E}_{(s,a)\sim \mathcal{D}}[L_2^\tau (Q_{\bar{\varphi}}(s,a) - V_\psi(s))],
    \label{eq:v_learning}
\end{align}
where $L_2^\tau(u)=|\tau - \mathbbm{1}(u<0)|u^2$ represents an expectile regression loss. The value network $V_\psi(s)$ is utilized in the learning of Q-values:
{\small
\begin{align}
    \mathcal{L}_Q(\varphi) = \mathbb{E}_{(s_t,a_t,s_{t+1})\sim\mathcal{D}}[r(s_t,a_t)+\gamma V_\psi(s_{t+1}) - Q_\varphi(s_t,a_t))^2].
    \label{eq:q_loss}
\end{align}
}

\vspace{-1mm}
\noindent{In the context of the relationship between these offline RL methods and our LBM, the policy $\pi(a|s)$ can be viewed as an expectation of the CoT $c_t$, expressed as}
\begin{align}
    \pi(a|s)= \mathbb{E}_{c_t\sim\pi(c_t|s)}\pi(a|s,c_t).
\end{align}
We denote the value of a state-action pair conditioned on the CoT $c_t$ as $Q(s_t, \tilde{a}_t)$. Our objective is to obtain an improved policy through an enhanced CoT by
\begin{align}
    \pi(a|s,c_t^{j}), c_t^{j}=\arg\max_{c_t}[Q(s_t, \tilde{a}_t) - Q(s_t,a_t)].
    \label{eq:cot_goal}
\end{align}
% which leads to our hierarchical LBM policy.
% If we regard the CoT $c_t$ as an action of the LBM-Think policy, 
For a practical implementation, we can assess the impact of CoT on decision-making by examining the difference in predicted Q-values for actions generated with and without CoT, termed \textit{relative-Q}. Specifically, we let LBM-Think generate a CoT $c_t$ using Eq.~(\ref{eq:lbm-think}) and send it to the trained LBM-Act to obtain an action $\tilde{a}_t$ via Eq.~(\ref{eq:lbm-act}). We then compare this with the dataset action $a_t$, which represents the ideal generation of a DT method without CoT. The \textbf{relative-Q} is calculated as:
{\small
\begin{align}
    \Delta Q = Q_\varphi(s_t, \tilde{a}_t) - Q_\varphi(s_t, a_t).
\end{align}
}
If $\Delta Q > 0$, it indicates that CoT has a positive impact on decision-making; otherwise, it suggests a negative impact.

Therefore, to achieve an expected objective in Eq. (\ref{eq:cot_goal}), we propose a stable and flexible method, termed Group relative-Q Policy Optimization (GQPO). 
Grasping the idea behind GRPO in finetuning an LLM, it is to drive the policy from $\pi_{old}$ to the near region of optimal policy $\pi^*$.
We can generate a group of CoTs $\{c_t^i\}_{i=1}^N$ and then assess their values via relative-Q, and then we find the best one $c_t^{j}$ that satisfies 
\begin{align}
    j = argmax_i \Delta Q_i,\quad s.t. \quad\Delta Q_i >0.
\end{align}
% Then, we let the model to do SFT on the selected CoTs.
Inspired by the AWR method in Eq.~(\ref{eq:awr}), we can interpret the CoT $c_t$ as the action and $\Delta Q$ as the advantage. Given that we can sample multiple CoTs simultaneously, the regression would be effective and stable when performed directly on the CoT $c_t^j$ with the highest relative-Q. This leads us to the objective of LBM-Think:
{\small
\begin{align}
    \mathcal{J}_{\text{GQPO}}(\theta)&=\mathbb{E}_{s,a\sim\mathcal{D},c_t^i\sim h_\theta}[\exp(\beta\Delta Q_i)\log h_\theta(c_t|s)] \notag \\
    &\propto \mathbb{E}_{s,a\sim\mathcal{D}, c_t^j\sim\{c_t^i\}_{i=1}^{N}}\log h_\theta(c_t^j|s).
\end{align}
}

After training the LBM-Think, we now give a detailed inference procedure of the LBM for auto-bidding in Fig. \ref{fig:inference}.

% This approach is flexible in incorporating expert knowledge and mitigating hallucination, as we can do filtering on the generations that satisfy the rules. 

% \begin{align}
%     \mathcal{J}_{GQPO} = 
% \end{align}

% This approach helps ensure more accurate Q-value estimation by addressing potential biases in the learning process.

\begin{table*}[t!]
\caption{Comparison between our LBM methods and previous non-LLM-based methods in AuctionNet.}
\vspace{-3mm}
\setlength\tabcolsep{8pt}
\renewcommand{\arraystretch}{1.0}
\resizebox{\linewidth}{!}{
\begin{tabular}{cc|cccccccccc}
\hline\hline
AuctionNet              &             & USCB & CQL  & IQL  & BCQ  & DT   & DT-Q & DiffBid & DiffBid-Q & LBM(P) & LBM(GQPO)     \\ \hline
\multirow{2}{*}{Dense}  & Conversions ($\uparrow$) & 267  & 336  & 322  & 321  & 364  & 371  & 316     & 319       & 376 $\pm${3.0}   & \textbf{382}$\pm${3.3}  \\
                        & Score ($\uparrow$)      & 157  & 171  & 281  & 270  & 329  & 334  & 214     & 260       & 320 $\pm${2.2}   & \textbf{348}$\pm${2.3}  \\ \hline
\multirow{2}{*}{Sparse} & Conversions ($\uparrow$)& 28.5 & 30.2 & 36.0 & 31.1 & 31.3 & 33.8 & 31.3    & 31.5      & 37.4 $\pm${0.3}  & \textbf{38.5} $\pm${0.3} \\
                        & Score ($\uparrow)$      & 17.5 & 22.2 & 30.0 & 30.5 & 29.6 & 31.1 & 23.1    & 25.1      & 32.9 $\pm${0.2}   & \textbf{33.4} $\pm${0.2} \\ \hline\hline
\end{tabular}}
\label{tab:comp_all}
\end{table*}

\begin{table*}[t!]
\centering
\caption{Comprehensive evaluation of LLM-based auto-bidding methods in AuctionNet.}
\vspace{-3mm}
\setlength\tabcolsep{8pt}
\renewcommand{\arraystretch}{1.0}
\resizebox{0.9\linewidth}{!}{
\begin{tabular}{cl|cccccccc}
\hline \hline
\multicolumn{1}{l}{}     & \multicolumn{1}{c|}{\multirow{2}{*}{Method}} & \multicolumn{2}{c}{\textbf{Budget Utilization} ($\uparrow$)} & \multicolumn{2}{c}{\textbf{CPA Ratio} ($\downarrow$)} & \multicolumn{2}{c}{\textbf{Conversions} ($\uparrow$)} & \multicolumn{2}{c}{\textbf{Score} ($\uparrow$)} \\
\multicolumn{1}{c}{}     & \multicolumn{1}{c|}{}                        & Dense             & Sparse            & Dense         & Sparse        & Dense          & Sparse         & Dense       & Sparse      \\ \hline
\multirow{6}{*}{LLM}     & Prompting                                    &     0.743              &   0.722                &     \textbf{0.878}          &   0.903            &          289      &     24.2           &   286          &    23.4         \\
                         & SFT                                          &   0.765                &    0.741               &          0.882      &       \textbf{0.804}        &          297      &     27.0           &   286          &   26.9          \\
                        & GRPO                                          & 0.860                 &     0.773              &    0.945           &      0.816         &    327            &  29.9              &     292        & 29.7            \\
                         & LLM-DT                                       & 0.902                  & 0.861                  &  0.936             & 0.847               & 366               & 36.0               & 328            & 32.5            \\ 
                         & Prompt-LLM-DT                                       &  0.914                 & 0.970                  &  0.931             &  0.965             & 355               & 35.8               &  322           & 30.3            \\ \cline{3-10} 
                         & -$\textit{ours}$              &                   &                   &               &               &                &                &             &             \\
                         & LBM(P)                        &  \textbf{0.981}                 & 0.956                  &  1.02             & 0.901              & 376               & 37.4               & 320            & 32.9            \\
                         & LBM(GQPO)                             & 0.938                  &  \textbf{0.974}                 &     0.96          &  0.901             &     \textbf{382}           & \textbf{38.5}               &          \textbf{348}   &  \textbf{33.4}           \\ \hline\hline
\end{tabular}}
\label{tab:gen_baseline}
\end{table*}

\section{Experiment}
\subsection{Setup}
% \vspace{-2mm}
\noindent\textbf{Datasets.} 
To evaluate the performance of bidding strategy decision-making in large-scale ad auctions, we employ AuctionNet and its sparse variant, a publicly available benchmark from Alibaba designed for ad auctions, based on a real-world online advertising platform \citep{auctionNet}. 
Each dataset comprises 21 advertising delivery periods, with each period containing approximately 5,000,000 impression opportunities, divided into 48 intervals. 
% The auto-bidding strategy adopted by the 
% A total of 48 advertisers of different settings participate in bidding, competing for every impression. 
% The auto-bidding strategy is iteratively adopted by each advertiser in every period to evaluate its performance under diverse advertiser settings.
Details are in Appendix \ref{app:exp}.

% {More details are in Appendix \ref{parameter_setting}.}

\noindent\textbf{Evaluation Metrics.}
To assess performance, we evaluate both the baselines and our method using four metrics:
% \vspace{-3mm}
\begin{itemize}
    \item  \text{\textbf{Conversions}}: in the maximize conversions bidding (MCB) task\footnote{https://support.google.com/google-ads/answer/7381968?hl=en}, advertisers care about the total received impression value, \textit{i.e.}, number of conventions $\sum_i o_i v_i$ in this task;
    \item \textbf{Budget Utilization}: ratio of the spent budget by the end of the period;
    \item \textbf{CPA Ratio}: advertisers are concerned with whether the strategy's realized CPA, $C_{real}$, is lower than the CPA constraint, $C$. Therefore, we use the CPA ratio, defined as $\text{ratio} =  {C_{real}}/{C}$, to assess this.
    \item \textbf{Score}: by introducing a 
    $
        penalty=min\{(\frac{1}{ratio})^2, 1\}$,
    we can get an additional metric as
    $
        score = (\sum_i o_i v_i) \times penalty.
    $
\end{itemize}
We use $\uparrow$ and $\downarrow$ to denote the direction of improved performance under these metrics and \textbf{bold} the best performance in results.

% \vspace{-2mm}
\noindent\textbf{Baselines.} We perform an extensive comparison of our approach against a diverse set of baselines, encompassing both non-LLM-based and LLM-based methods.
For non-LLM-based methods, our comparison includes RL techniques such as \textbf{USCB} \citep{USCB}, \textbf{BCQ} \citep{bcq}, and \textbf{CQL} \citep{cql}. We also evaluate decision transformer (DT)-based methods, including \textbf{DT} and its variant \textbf{DT-Q}, which utilizes GAS \cite{gas} for fine-tuning a DT with Q-values. Additionally, we consider diffuser-based methods like \textbf{Diffbid} \cite{aigb} and \textbf{Diffbid-Q}, which leverage CBD \cite{cbd} to align a diffuser with a trajectory-level reward model.
For LLM-based methods, we compare against a \textbf{Prompting} method, a \textbf{SFT} method, and a \textbf{GRPO} method within language modality. Furthermore, we evaluate \textbf{LLM-DT} and {Prompt-LLM-DT}, which handle numerical sequence inputs to produce numerical actions, with \textbf{Prompt-LLM-DT} incorporating an additional language task description.

% takes an extra language task description compared to LLM-DT.
% Our methods, denoted as \textbf{LBM(prompting)}, utilize a non-finetuned LLM alongside a trained LBM-Act, and \textbf{LBM(GQPO)}, which employs a finetuned LLM and a trained LBM-Act.

\noindent\textbf{Implementation Details.} 
The hyperparameters of baselines are set based on the default values referenced in the original papers, with further tuning performed to optimize performance.
The LBM-Think model can be based on any pretrained LLMs. In this work, we use Qwen2.5-3B-Instruct for the main results. To train LBM-Think, we utilize 2000 samples obtained through the GQPO method with $N=3$. We then fine-tune it with a batch size of 64 and a learning rate of 1e-6 for 5 epochs, using the Llama-factory framework under full parameters \cite{zheng2024llamafactory}.
The LBM-Act model is built on a smaller LLM, specifically Qwen2.5-0.5B-Instruct \cite{qwen}. 
We employ a three-layer MLP with dimensions [896, 896, 896] to process items in the numerical sequence into embeddings that match the size of the token embeddings. The output from the final transformer layer is projected into actions using another three-layer MLP with dimensions [896, 896, 1].
The training of LBM-Act undergoes a total of 400,000 training steps using the AdamW \cite{adamw} optimizer with a learning rate of 1e-5 and a batch size of 64. 
The training process is conducted using the PyTorch framework on 8 GPUs.
% NVIDIA H800 GPUs.
The results presented are averaged over five random runs.

\textbf{Details of the baselines and implementations can be seen in Appendix \ref{app:baseline}}.

\vspace{-4mm}
\subsection{Comparison to Auto-Bidding Baselines}
\vspace{-1mm}
% To evaluate the potential of LLMs as a superior backbone for auto-bidding, 
We start by comparing our LBM method with established auto-bidding approaches, including offline RL methods and various generative backbones, using Conversions and Scores as metrics.
As shown in Table \ref{tab:comp_all}, the generative method based on DT exhibits better performance than previous offline RL methods, demonstrating the efficiency of generative models in learning a policy from a large-scale dataset. Without incorporating Q-value, our LBM method variant, referred to as LBM-pretrained (denoted as \textbf{LBM(P)} in the results), which employs a pretrained LLM as the LBM-Think model alongside a trained LBM-Act model, could generally outperform previous methods, particularly DT. Even with various budget settings in the test, our LBM method still demonstrates generalizability and consistently outperforms DT, as shown in Table \ref{tab:budget}.
When incorporating Q-value to implement the GQPO method, a fine-tuned LBM-Think integrated into the LBM framework (denoted as \textbf{LBM(GQPO)}) further enhances the performance of LBM(P), which demonstrates the advantage of finetuning for LLMs.
In summary, the performance comparison across various methods highlights our LBM as a superior generative backbone for auto-bidding.

One advantage of our LBM method is its ability to leverage the prior knowledge embedded in LLMs, which can be more efficient than relying solely on reward design for training like DT. We utilize the Score metric as an evaluation tool for this purpose. While the current reward design focuses only on conversions, the Score metric offers a more comprehensive evaluation by considering both conversions and the CPA ratio through a penalty term. However, this penalty term is calculated over the entire bidding period, making it challenging to assign accurately to a single-step signal, thus complicating the return-to-go setting in DT methods.
Empirically, we propose a more sophisticated return-to-go design for DT to achieve a better Score, expressed as 
$
rtg_w = \sum_{i=t}^{T} r_i + w \times penalty_{t:T}
$
where \( penalty_{t:T} \) is calculated for future timesteps in the training stage and set to 1 during the inference stage. The parameter \( w \) serves as a reweighting factor to balance conversions and the CPA ratio. Intuitively, setting \( w = 0 \) encourages the DT to prioritize conversions, while \( w > 0 \) shifts focus towards the CPA ratio. By adjusting this reweighting factor, we can balance conversions and the CPA ratio to achieve a better Score.
As demonstrated in Table \ref{tab:reward_set}, DT with \( w = 0.2 \) achieves the highest score. However, the LLM-based method can bypass this inefficient training based on reward design due to its inherent language knowledge and generalization capabilities, allowing it to outperform them.

\begin{table}[h]
\caption{Performance with various reward setting.}
\vspace{-3mm}
\resizebox{0.47\textwidth}{!}{
\begin{tabular}{l|cccccc}
\hline\hline
                  & \multicolumn{5}{c}{\textbf{DT-score}} & \textbf{LBM} \\
{Reweight $w$}    & 0  & 0.1  & 0.2  & 0.5  & 1.0   &                     \\ \hline
Budget Utlization ($\uparrow$) & 0.884    & 0.839     & 0.818     & 0.780     & \textbf{0.632}           & 0.938                    \\
CPA Ratio ($\downarrow$)        & 0.895   & 0.846     & 0.826     & 0.831     & 0.725           & \textbf{0.960}                    \\
Conversions ($\uparrow$)      & 364   & 354     & 352     & 329     & 278           &\textbf{382}                     \\
Score ($\uparrow$)            & 329   & 340     & 343     & 323     & 276          &\textbf{348}                     \\ \hline\hline
\end{tabular}
}
\label{tab:reward_set}
\vspace{-3mm}
\end{table}

\begin{figure}[htbp]
    \centering
    \begin{subfigure}{0.15\textwidth}
        \centering
        \includegraphics[width=\textwidth]{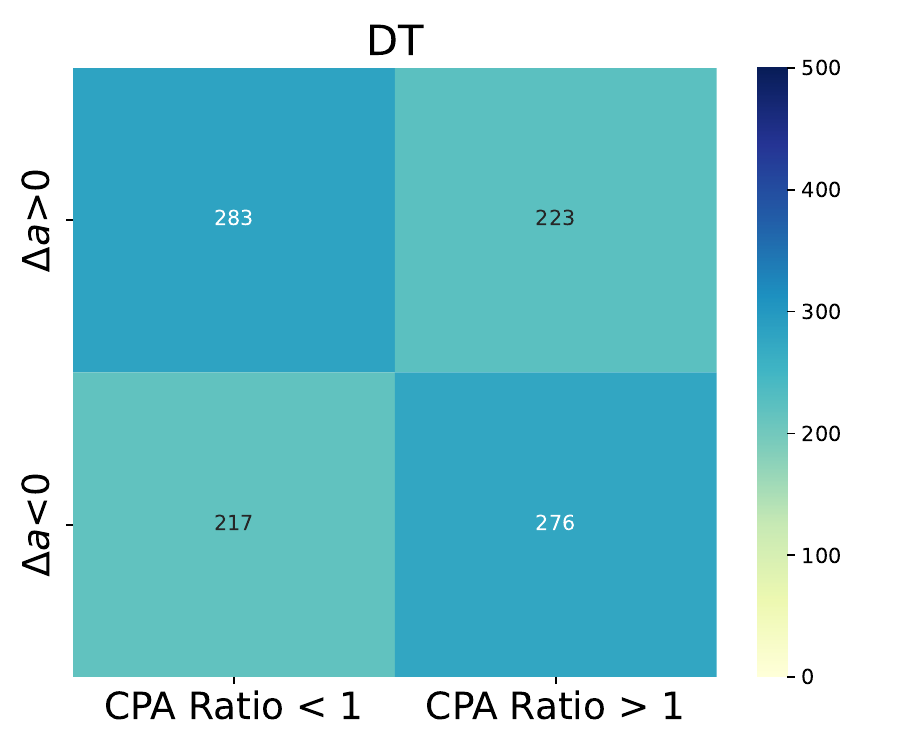}
        \caption{DT}
    \end{subfigure}
    \begin{subfigure}{0.15\textwidth}
        \centering
        \includegraphics[width=\textwidth]{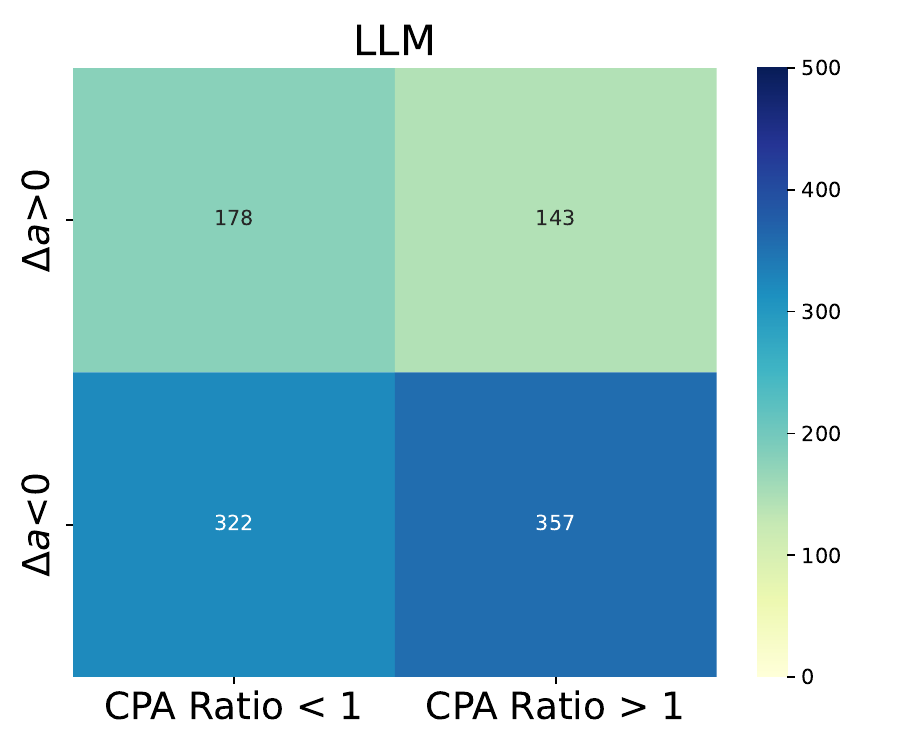}
        \caption{LLM}
    \end{subfigure}
    \begin{subfigure}{0.15\textwidth}
        \centering
        \includegraphics[width=\textwidth]{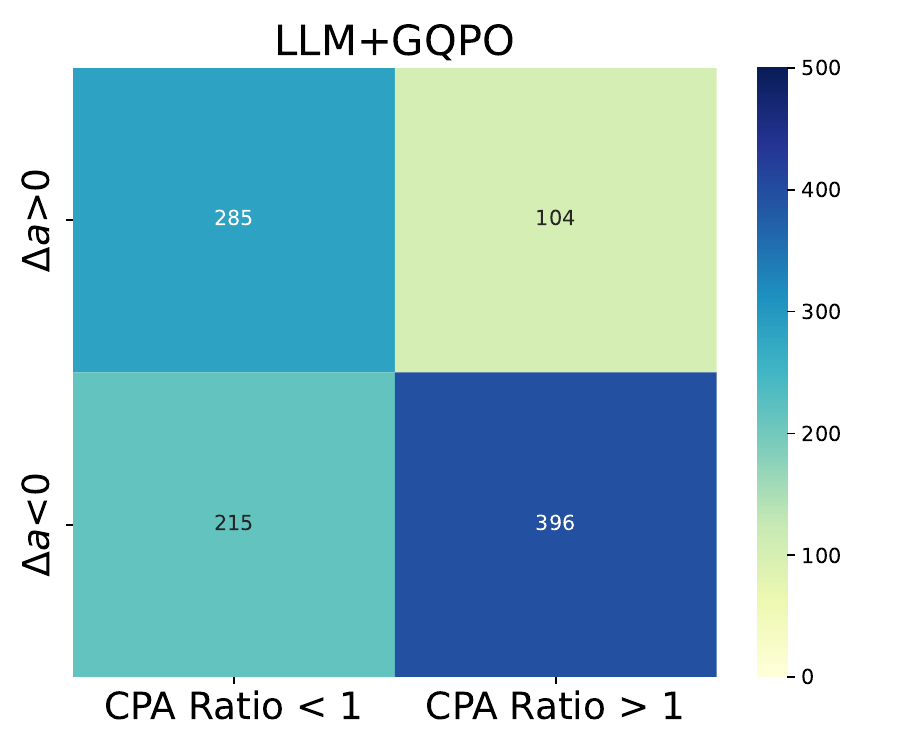}
        \caption{LLM+GQPO}
    \end{subfigure}
    \vspace{-3mm}
    \caption{Relationship between the CPA ratio and model behavior by 1000 random samples. When the CPA ratio exceeds 1, it indicates that the CPA constraint is not satisfied, suggesting a high likelihood of decreasing the bidding parameter, \textit{i.e.}, $\Delta a<0$. When the CPA ratio is smaller than 1, it will be more likely to increase the bidding parameter $\Delta a>0$. However, this relationship is not clearly reflected in DT and pretrained LLM, but is more evident in LLM finetuned with GQPO.}
    \label{fig:relation}
\end{figure}

The advantage of prior knowledge could also be illustrated in Fig. \ref{fig:relation}. A fundamental principle in auto-bidding is to adhere to economic constraints while maximizing conversions. For instance, if the CPA ratio exceeds 1, it is more reasonable and likely to decrease the bidding parameters to meet the constraint. Conversely, when the CPA ratio is below 1, the model can safely increase the bidding parameters to gain more impression opportunities. As shown in Fig. \ref{fig:relation}, where we randomly sampled 1000 samples, the behavior of the LLM finetuned with GQPO clearly aligns with this principle, whereas DT models show weaker adherence, underscoring the crucial role of reasoning with prior knowledge in auto-bidding.

% 

% \newpage
\begin{table}[h]
\caption{Language instruction's impact on performance.}
\vspace{-3mm}
\resizebox{0.47\textwidth}{!}{
\begin{tabular}{l|cccc}
\hline\hline
    Instruction       & Budget Utilization($\uparrow$) & CPA Ratio($\downarrow$) & Conversions($\uparrow$) & Score($\uparrow$) \\ \hline
Increase   & \textbf{0.969}                  &  1.028         & 368            & 315      \\
Decrease   & 0.894                   & \textbf{0.924}          & 373            & 325      \\
base & 0.938                   &  0.960         &  \textbf{382}           & \textbf{348}      \\ 
\hline\hline
\end{tabular}}
\label{tab:instruction}
\end{table}

% \usepackage{multirow}
\begin{table}[ht]
\caption{Generalization to various budget settings. Performance under conversions.}
\vspace{-3mm}
\begin{tabular}{ll|ccccc}
\hline\hline
                        & Budget Ratio & 0.5 & 0.75 & 1.0 & 1.25 & 1.5 \\ \hline
\multirow{2}{*}{Dense}  & DT           & 186    & 234     & 364    & 393     & 445    \\
                        & LBM(P)   &\textbf{198}     & \textbf{285}     & \textbf{376}    & \textbf{462}     & \textbf{520}    \\ \hline
\multirow{2}{*}{Sparse} & DT           & 18.2    & 27.5     & 31.3    & 43.5     & 50.0    \\
                        & LBM(P)   & \textbf{21.3}    & \textbf{28.5}     & \textbf{37.4}    &  \textbf{43.6}    & \textbf{50.6}    \\ \hline\hline
\end{tabular}
\label{tab:budget}
\vspace{-3mm}
\end{table}

\vspace{-3mm}
\subsection{Performance of LLM-based Methods}
\vspace{-1mm}
To apply the LLM for auto-bidding, we initially explored multiple methods within the pure language modality, where the long numerical sequence information is directly translated into language tokens, and the generated action is also represented by language tokens. First, through \textbf{Prompting}, we use prompt engineering to apply an LLM to generate the CoT and an action in language format without fine-tuning, which can be seen as the RTB-agent method \cite{rtb-agent}. Second, with \textbf{SFT}, we perform supervised fine-tuning to enable the LLM to generate an action directly in language format. Lastly, using \textbf{GRPO}, we fine-tune the LLM to generate CoT and action by using the GRPO method to stimulate its reasoning ability, where the reward is the L1 distance between the generated action and the dataset action.
However, as shown in Table \ref{tab:gen_baseline}, these three methods perform poorly in auto-bidding that cannot fully utilize the budget, potentially due to the inefficient representation of long sequential state information. 
Additionally, we observed that GRPO's training is unstable in this task, often resulting in shorter or absent CoT, as shown in { Fig. \ref{fig:train_response_length}}.
This observation also motivated us to propose a hierarchical structure for an LLM-based method by decoupling reasoning and acting into separate models.

\begin{figure}
    \centering
    \includegraphics[width=0.9\linewidth]{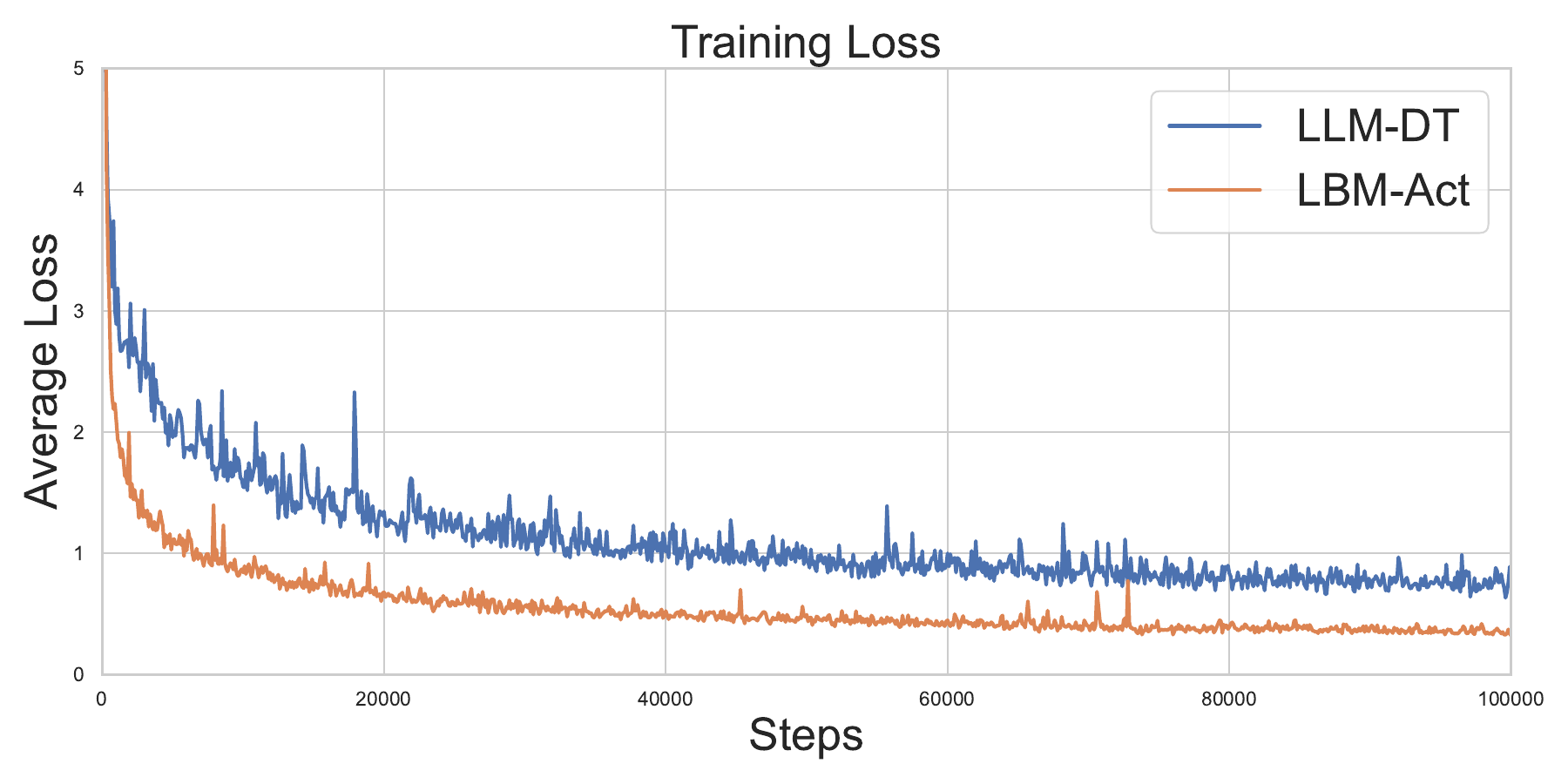}
    \vspace{-5mm}
    \caption{Training loss curve of LLM-DT and LBM-Act, supporting the efficiency of language-guided learning.}
    \label{fig:loss}
\end{figure}

% [efficient training, aware of language]
Therefore, we tend to the LLM-DT method, which employs the LLM for decision-making tasks in a continuous state and action space.
As it does not contain information from the language, we add a variant to LLM-DT by incorporating the task description in the input embedding, termed Prompt-LLM-DT, making the LLM understand the basic task objective.
These LLM-DT methods show a better performance than the above three LLM-based methods in single language modality, demonstrating the efficiency in representing the numerical information via additional embedding layers.
However, when compared to our LBM methods, their performance is limited, underscoring the advantages of incorporating reasoning by LBM-Think. 
This can also be attributed to the efficiency of LBM-Act in integrating the two modalities and utilizing the language reasoning component for decision-making. 
% Additionally, the LLM-DT is essentially a DT initialized with LLMs, which remains constrained by the inefficient training from reward signals. 
As demonstrated in the visualizations in Fig. \ref{fig:attention} and Fig. \ref{fig:loss}, LBM-Act exhibits faster convergence of the training loss compared to LLM-DT and provides informative attention scores.
We also included an instruction-following experiment in Table \ref{tab:instruction}, where the reasoning component is instructed to either increase or decrease the bidding parameter. The results demonstrate strong instruction-following capabilities, as a decreased bidding parameter leads to lower budget utilization and a lower CPA ratio. This highlights the flexibility of using language instructions to modify auto-bidding strategies.
% without the need for fine-tuning.

\begin{table}[]
\caption{Impact of LBM-Think's model size.}
\vspace{-3mm}
\resizebox{0.45\textwidth}{!}{
\begin{tabular}{c|cccc}
\hline\hline
\multicolumn{1}{l|}{}           & \multicolumn{2}{c}{Dense} & \multicolumn{2}{c}{Sparse} \\
\multicolumn{1}{l|}{Model Size} & Conversions    & Score    & Conversions     & Score    \\ \hline
3B                              & 376               & 320         &  37.4               & 32.9         \\
7B                              & 375               & 320         & 37.3               & 34.6         \\
32B                             & 374               & 318         &  37.1               & 33.5         \\ \hline\hline
\end{tabular}}
\label{tab:model_size}
\vspace{-3mm}
\end{table}

% \noindent
\vspace{1mm}
\noindent{\textbf{Ablation Study.}}
Besides the LLM-DT and Prompt-LLM-DT supporting the importance of the reasoning ability of LBM-Think for auto-bidding, we further evaluate the performance of different LLM choices as the backbone for LBM-Think, including Qwen2.5-3B-Instruct, Qwen2.5-7B-Instruct, and Qwen2.5-32B-Instruct.
As the results shown in Table \ref{tab:model_size}, the performance is very close, showing the robustness of the LBM method, and a 3B LLM could be sufficient in handling the auto-bidding task.

\vspace{-1mm}
\section{Related Works}
% \vspace{-1mm}
\textbf{Auto-Bidding Methods.}
 Initially, traditional control methods like PID \cite{pid} were employed to optimize budget spending using predefined curves, while OnlineLP \cite{onlineLP} adjusted bids based on traffic forecasts. As online bidding environments grew more complex, RL algorithms such as USCB \cite{USCB}, SORL \cite{sorl}, and MAAB \cite{maab} became essential for decision-making. Offline RL methods, which learn policies from existing datasets without online interaction, have been particularly successful. Key methods include BCQ \cite{bcq}, which restricts the action space; CQL \cite{cql}, which regularizes Q-values; and IQL \cite{iql}, which enables multi-step updates without querying out-of-sample Q-values.
 However, these methods are limited by the MDP assumption, whereas generative models offer greater potential for auto-bidding. Recently, DiffBid \citep{aigb}, an innovative generative auto-bidding method, has shown the remarkable potential of generative models over RL by directly using a decision diffuser for planning and control, although it has not yet been tested in large-scale auctions.
GAS \citep{gas} and GAVE \citep{gave} propose data augmentation techniques to enhance the quality of offline datasets, but they still rely on the single-step reward.
RTB-agent \cite{rtb-agent} proposes a prompting method based on LLMs for auto-bidding, which we found is limited in large-scale auctions.

% \noindent\textbf{Non-LLM-based Methods for Auto-Bidding.}

\noindent\textbf{LLM for Decision-Making.}
LLMs have recently been explored as autonomous decision-making agents across domains such as navigate web pages \cite{zhou2023webarena},
android device control \cite{rawles2023androidinthewild} and robotics \cite{pertsch2025fast}.  Early studies have proposed structured
workflows that integrate reasoning, action, and reflection steps such as ReAct \cite{yao2023react} and Reflexion \cite{shinn2023reflexion}, as well as interaction through memory \cite{xu2025mem}, retrieval systems \cite{gao2023retrieval}, and external tools \cite{wang2024mobile}. Recent research has shifted beyond static prompting toward adaptive learning via supervised fine-tuning (SFT) with limited human demonstrations \cite{prabhakar2025apigen}. To further allow LLMs to operate in dynamic environments, RL has become essential for enhancing decision-making capabilities of LLMs. PPO \cite{ppo} and its variant GRPO \cite{guo2025deepseek} are the most widely adopted. Building on this line of work, GiGPO \cite{gigpo} introduces fine-grained credit assignment by leveraging repeated environment states. However, on-policy policy gradient methods often face limitations in real-world agent tasks, where interacting with the actual environment is expensive and inefficient \cite{luo2025mcp}. To overcome this, offline RL approaches have emerged to leverage static experience data. Digi-Q \cite{digi-q} trains VLM-based Q-functions to filter offline action data.
LLM-DT \cite{llm-dt, llm-dt-1} finetunes an LLM in a DT manner.
However, these methods do not focus on enhancing reasoning ability.

% derive agent policies without online rollouts, while OREO \cite{wang2024offline} fine-tunes LLMs by enforcing soft Bellman consistency through joint value and policy optimization.

\vspace{-2mm}
\section{Limitations and Conclusion}
% \vspace{-1mm}
% There are some limitations of this work. 
Due to safety concerns, we focus on offline fine-tuning techniques rather than continual training of LLMs via real-world rollouts. As a result, the extent of performance improvement remains limited.
Therefore, a promising future direction is to fine-tune LLMs in real advertising environments with safety control.
Additionally, the inference latency could be explored by incorporating more advanced acceleration methods, while we only use vLLM \cite{vllm} for acceleration in this work. 
Therefore, our method may face limitations in auto-bidding services that require extremely high-frequency adjustments. However, we emphasize that a common setting involves much longer intervals, which can be as long as 30 minutes, thus providing ample time for LLM-based methods.
A discussion and experiment on inference latency can be seen in Appendix \ref{app:latency}

% As some auto-bidding scenarios could have limited resources, there 

The increasing complexity of ad auctions necessitates advanced auto-bidding strategies, while current methods using offline reinforcement learning or generative approaches face challenges due to their black-box nature and direct applying LLMs is hindered by the need for precise actions and specialized bidding knowledge.
In this work, we introduce a hierarchical Large auto-Bidding Model (LBM), comprising LBM-Think for reasoning and LBM-Act for action generation. Our approach utilizes a dual embedding mechanism for efficient modality fusion and introduces GQPO, an offline reinforcement fine-tuning technique, to enhance decision-making without relying on rollouts in the real world. Experiments demonstrate the LBM's excellence in training efficiency and generalization, setting the stage for future auto-bidding approaches based on LLMs.

\begin{acks}
This research is supported by the Ministry of Education, Singapore, under its Academic Research Fund Tier 1 (RG18/24).
\end{acks}

\balance
\bibliographystyle{ACM-Reference-Format}
\bibliography{sample-base}

% \newpage
% \clearpage
\appendix

\section{Appendix}
\subsection{Dataset Details} \label{app:exp}
The AuctionNet benchmark comprises two datasets: AuctionNet and AuctionNet-sparse. AuctionNet-sparse is a sparser version of AuctionNet, featuring fewer conversions. Each dataset includes 21 advertising delivery periods, with each period containing approximately 5,000,000 impression opportunities, divided into 48 intervals. 
The models are trained on these two datasets, while 5,000 randomly selected trajectories from each dataset are used exclusively for evaluating the visualization of the generation.
Detailed parameters are provided in Table \ref{tab:dataset_setting}.

\begin{figure*}[h!]
    \centering
    \includegraphics[width=\linewidth]{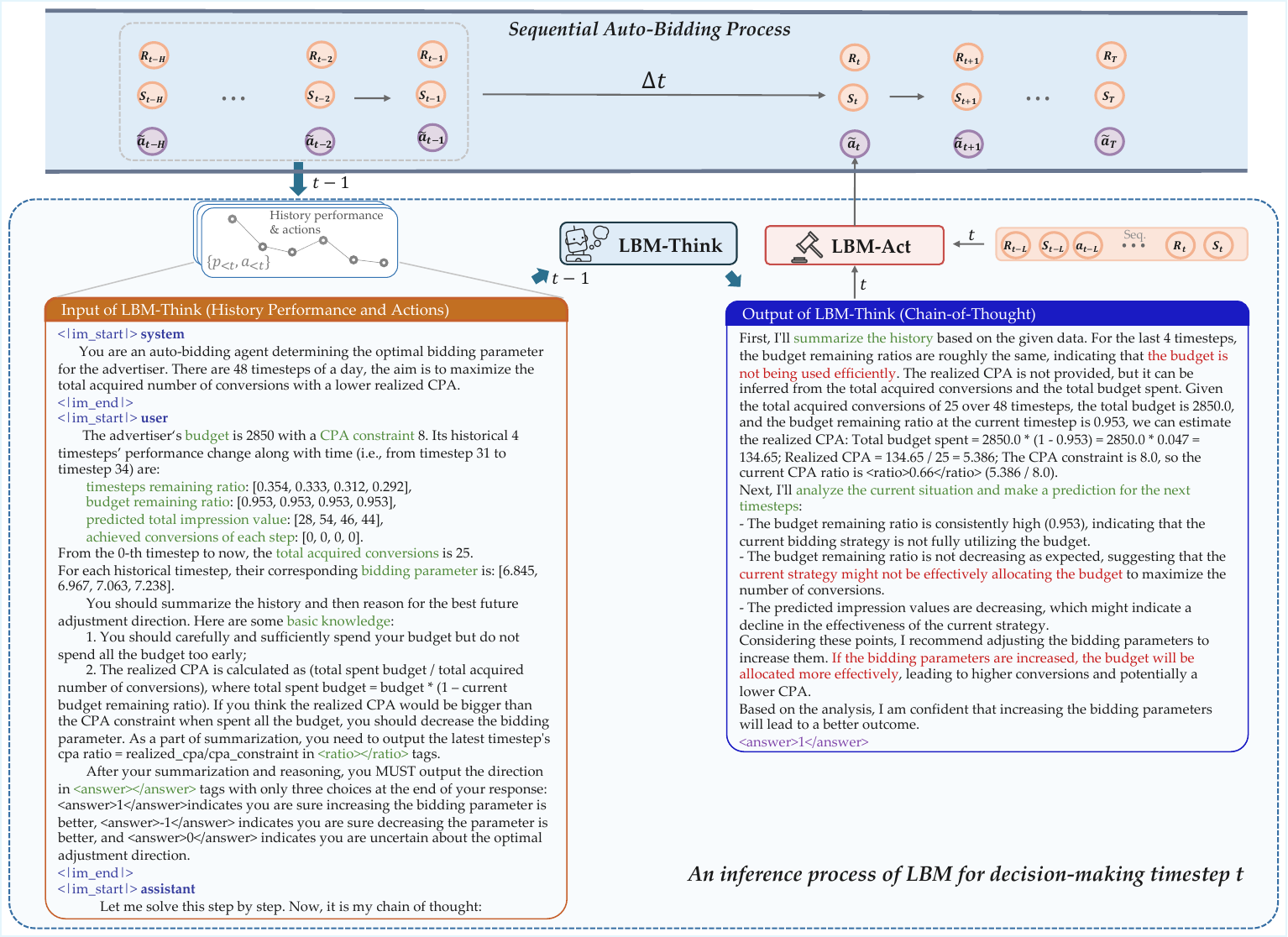}
    \caption{Illustration of the inference procedure of LBM with a detailed case of the prompt and CoT. For a timestep $t$, the LBM-Think first generates a CoT at timestep $t-1$ within the interval $\Delta t$. Then, when timestep $t$ arrives, the LBM-Act receives the pre-generated CoT along with the detailed numerical sequence information to generate an action, which involves adjusting the bidding parameter.}
    \label{fig:inference}
\end{figure*}

\subsection{Computational Efficiency}\label{app:latency}

We measured the inference latency using vLLM for acceleration on an H800 GPU for all models. As shown in Table \ref{tab:latency}, the LBM-Think based on the Qwen2.5-3B-Instruct model achieves the fastest speed, while the LBM-Act requires significantly less time.
As illustrated in Fig. \ref{fig:inference}, the LBM-Think has a maximum allowable inference latency, denoted as $\Delta t$, which is the interval between two consecutive decision-making timesteps.
Given that current industrial auto-bidding methods operate at a low frequency, sometimes as infrequently as every half hour, the LBM-Think of the current inference latency is able to be deployed.
However, we emphasize the importance of fully harnessing the potential of smaller LLMs, such as a 3B model, to achieve better computational efficiency.

\begin{figure*}[h!]
    \centering
    \begin{subfigure}[b]{0.33\textwidth}
        \includegraphics[width=\textwidth]{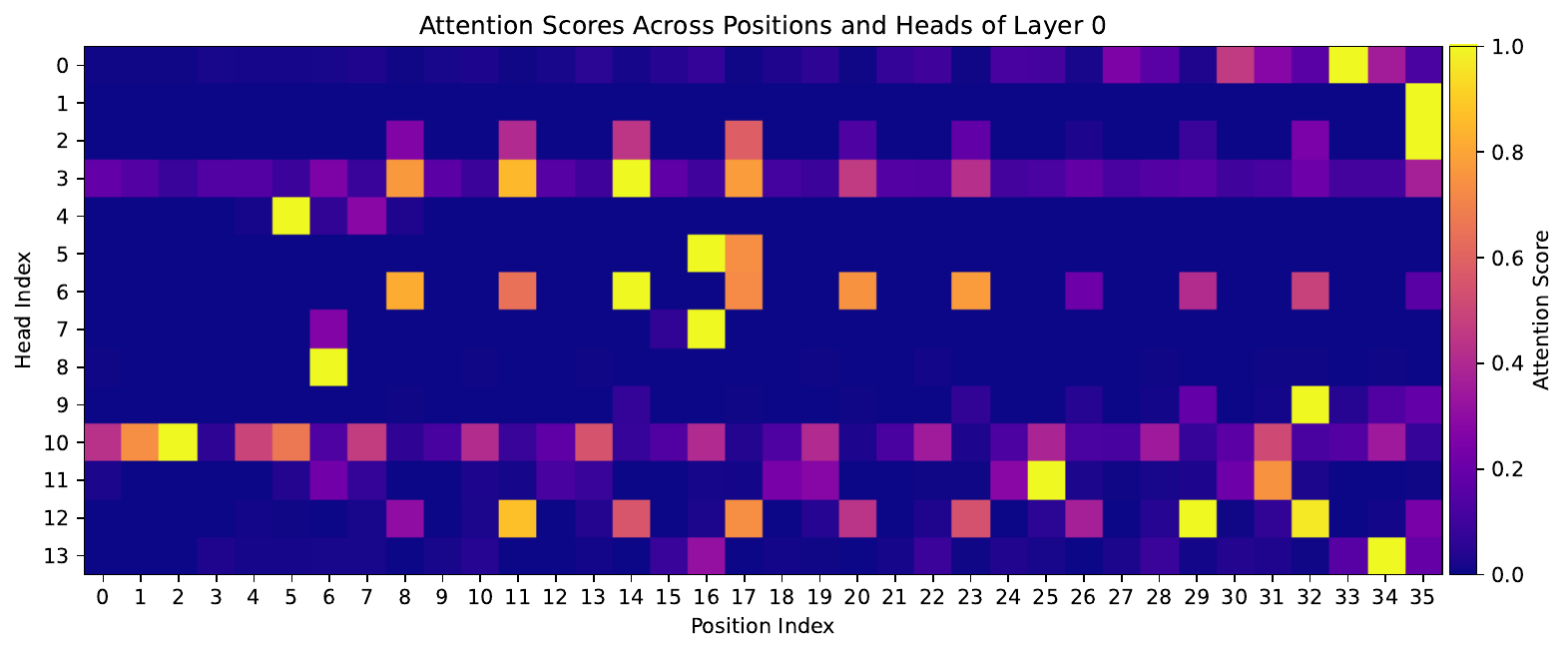}
        \vspace{-6mm}
        \caption{Attention scores of the beginning layer.}
        \label{fig:image1}
    \end{subfigure}
    % \hfill
    \begin{subfigure}[b]{0.33\textwidth}
        \includegraphics[width=\textwidth]{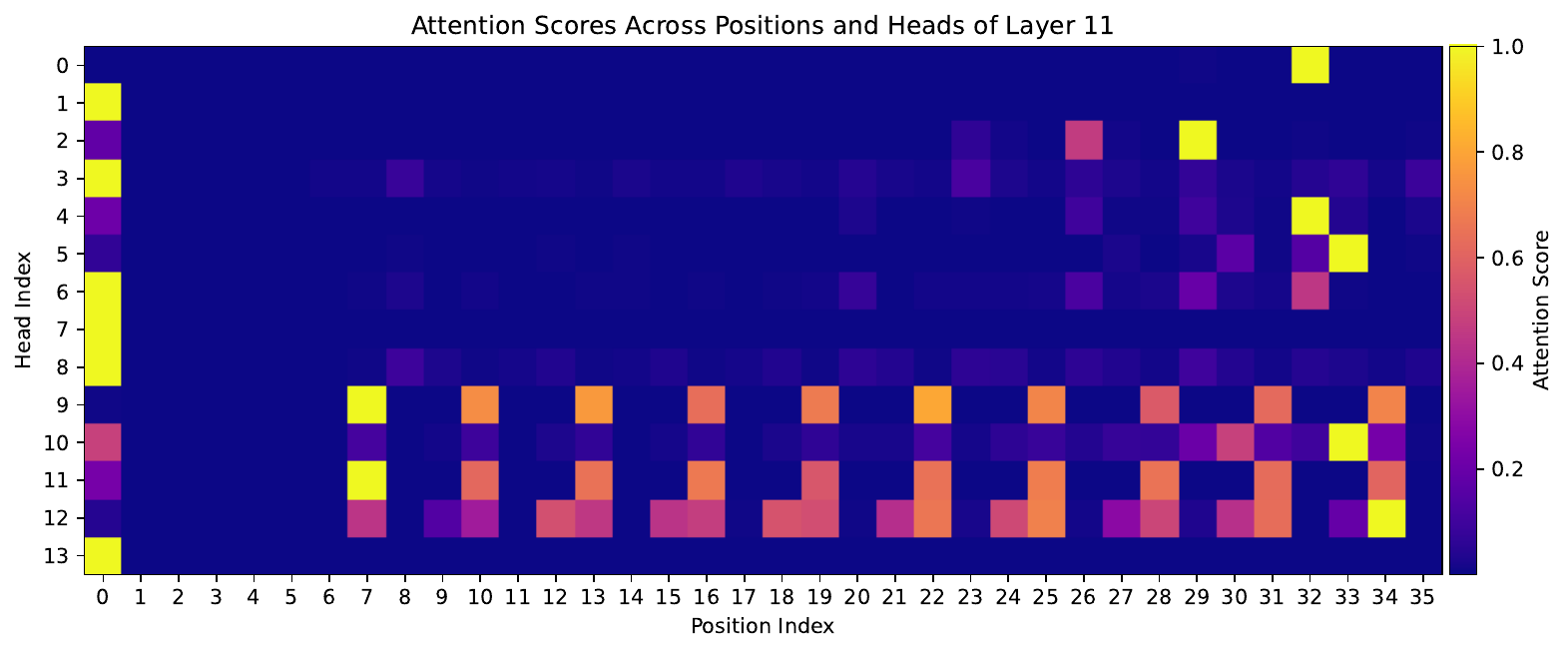}
        \vspace{-6mm}
        \caption{Attention scores of the middle layer.}
        \label{fig:image2}
    \end{subfigure}
    % \hfill
    \begin{subfigure}[b]{0.33\textwidth}
        \includegraphics[width=\textwidth]{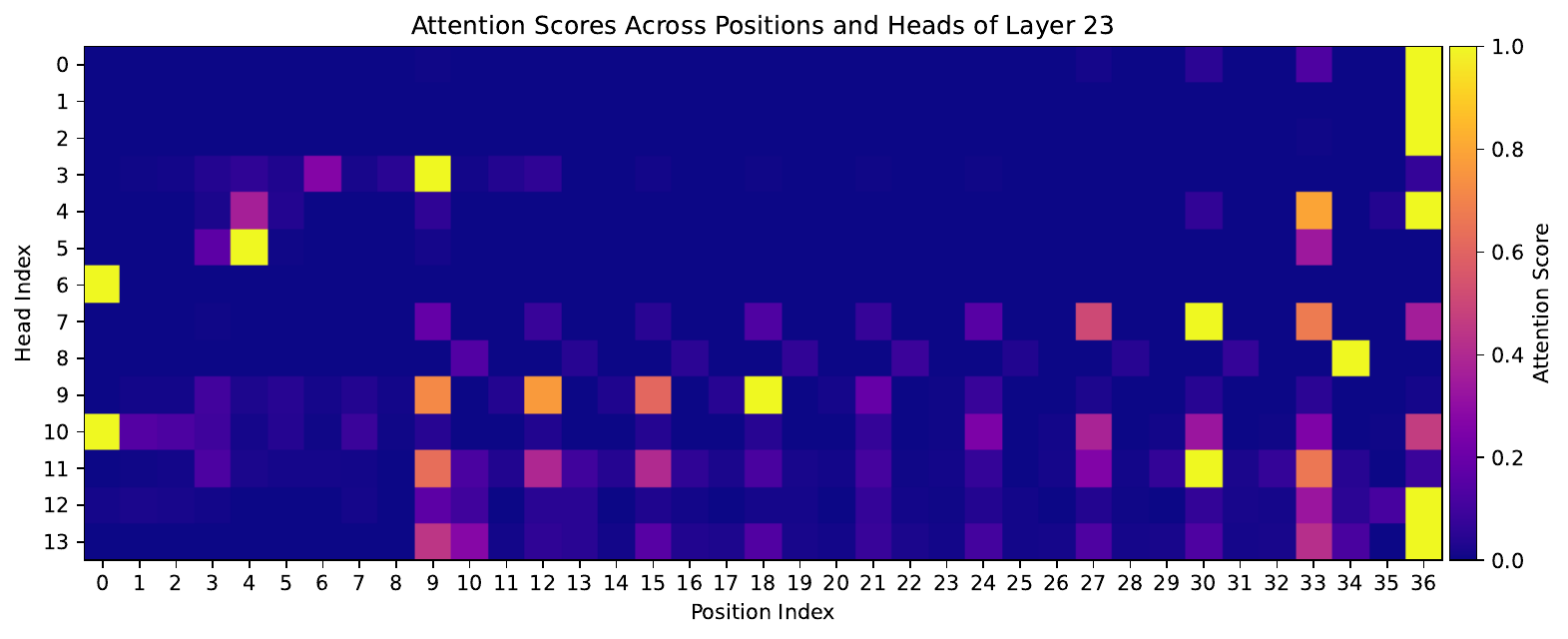}
        \vspace{-6mm}
        \caption{Attention scores of the final layer.}
        \label{fig:image3}
    \end{subfigure}
    \vspace{-6mm}
    \caption{Visualization of attention scores varying across different positions and heads. For simplicity, we let positions from \#0 to \#6 as the language part, while the remaining positions are the numerical part. The two modalities are effectively fused using different attention heads. For instance, heads such as \#4, \#5, \#6, and \#10 in the final layer predominantly focus on the language part, whereas other heads are more attuned to the numerical parts. Interestingly, we observed that the attention in the middle layers distinctly separates the language parts (positions 0-6) from the numerical parts, with position 7 marking the beginning of the numerical part.}
    \label{fig:attention}
\end{figure*}

\begin{table}[ht!]
  \centering
  \caption{The parameters of AuctionNet and AuctionNet-sparse.}
  \scalebox{0.9}{
    \begin{tabular}{ccc}
    \toprule
    Params & AuctionNet & AuctionNet-Sparse \\
    \midrule
    Trajectories & 479,376 & 479,376 \\
    Delivery Periods & 9,987 & 9,987 \\
    Time steps in a trajectory & 48    & 48 \\
    State dimension & 16    & 16 \\
    Action dimension & 1     & 1 \\
    % Return-To-Go Dimension & 1     & 1 \\
    Action range & [0, 493] & [0, 8178] \\
    Impression's value range & [0, 1] & [0, 1] \\
    CPA range & [6, 12] & [60, 130] \\
    Total conversion range & [0, 1512] & [0, 57] \\
    \bottomrule
    \end{tabular}%
    }
  \label{tab:dataset_setting}
\end{table}

\begin{table}[h!]
  \centering
  \caption{Inference Latency of different modules.}
  \scalebox{0.9}{
    \begin{tabular}{l|cccc}
    \toprule
                      & \multicolumn{3}{c}{\textbf{LBM-Think}} & \textbf{LBM-Act} \\
    \#params           & 3B          & 7B          & 32B        & 0.5B             \\ \midrule
    Inference Latency & 2.5s        & 3.6s        & 8.9s       & 63ms             \\ \bottomrule
    \end{tabular}}
\label{tab:latency}
\end{table}

\begin{figure}[]
    \centering
    \includegraphics[width=0.4\textwidth]{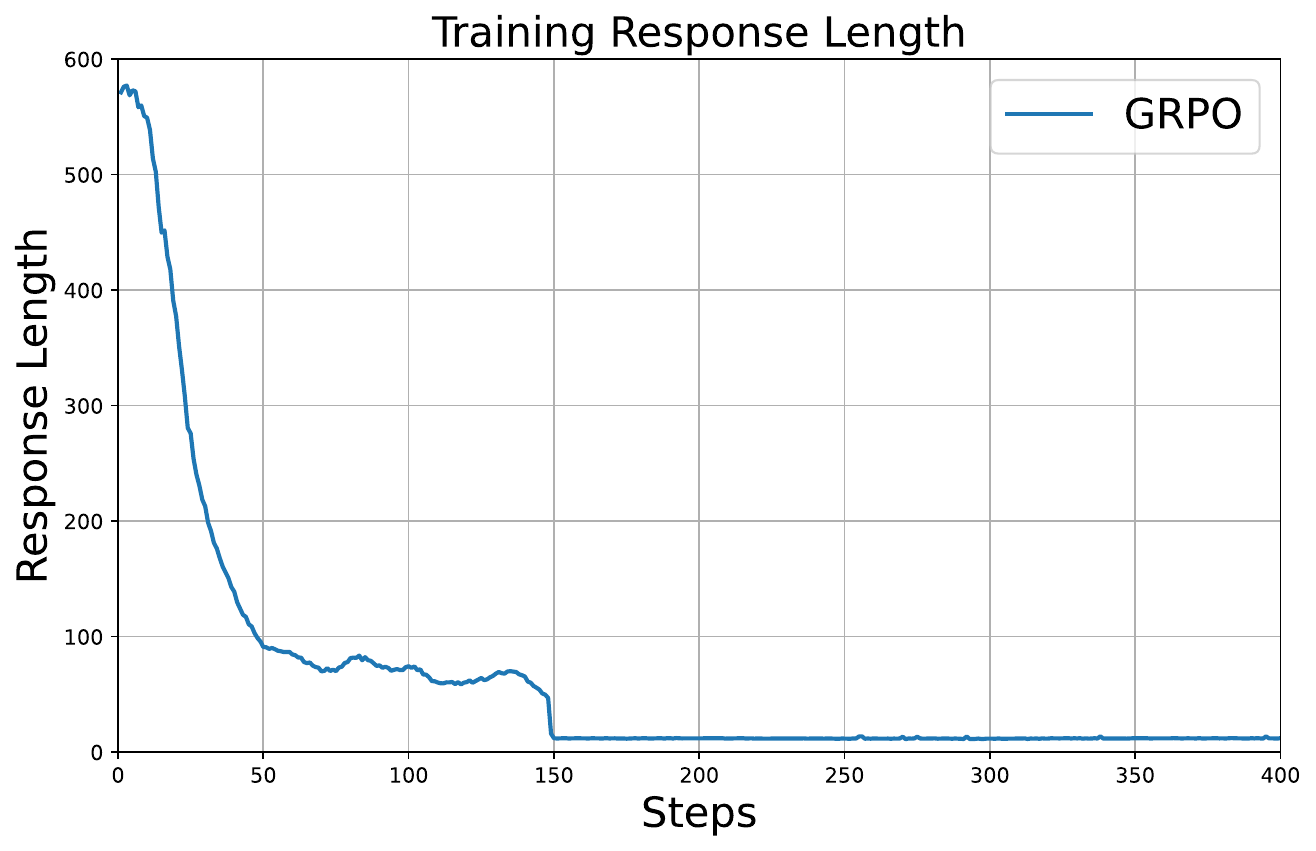}
    \caption{Finetune the LLM using GRPO so that it first performs step-by-step reasoning and then outputs an action. However, after around 150 training steps, the model only outputs the action without performing additional reasoning.}
    % \Description{Training response length of the LLM, showing reasoning steps decrease after 150 steps.}
    \label{fig:train_response_length}
\end{figure}

\subsection{Details of Baselines and Implementation}\label{app:baseline}
% \textbf{LLM-based Baselines.}
\textbf{Prompting} and \textbf{SFT} are two straightforward techniques for processing prompts that include task descriptions and observed sequential information, such as states, previous actions, and return-to-go, to generate actions in the language modality as responses. The \textbf{Prompting} method relies solely on prompt engineering, crafting prompts that encapsulate task and sequence details and instruct the model to perform reasoning to elicit desired outputs. In contrast, \textbf{SFT} employs a supervised fine-tuning approach, training the model on labeled action data in the language modality to enhance its ability to interpret prompts and produce accurate actions, akin to a Decision Transformer (DT) training manner but in the language modality. 
As the LLM with SFT has no CoT, we further employ the \textbf{GRPO} to fine-tune the LLM for generating CoT and actions, aiming to enhance its reasoning capabilities while keeping action generation ability in a single LLM. 
Nonetheless, as illustrated in Table \ref{tab:gen_baseline}, this GRPO approach underperforms in auto-bidding, failing to fully utilize the budget. Furthermore, we noticed instability in GRPO's training for this task, frequently leading to shorter or missing CoT, as depicted in Fig. \ref{fig:train_response_length}.
To achieve computational efficiency, all three methods are built on Qwen2.5-3B-Instruct.

\textbf{LLM-DT} is specifically designed to deploy LLMs for decision-making in continuous action spaces. Its inputs and outputs are the same as those of a Decision Transformer (DT); specifically, the input is a sequence containing return-to-go, states, and actions in numerical modality, and the output is an action in numerical modality. 
The key distinction between LLM-DT and DT is that LLM-DT utilizes transformer layers directly from a pretrained LLM. Since these transformer layers are trained to process embeddings of language tokens, LLM-DT introduces additional embedding layers to handle numerical sequences. This is achieved using a three-layer multilayer perceptron (MLP), which is the same as our decision embedding layer. In this way, a sequence of $n$ items can be projected into $n$ embeddings of the same size as the token embeddings, providing a highly informative representation, unlike representing a number with multiple token embeddings.
LLM-DT's training process is identical to that of a Decision Transformer (DT). 
\textbf{Prompt-LLM-DT} extends LLM-DT by incorporating a task description in natural language to determine if it can enhance decision-making. This language component is encoded using the pretrained token embedding layers of the LLM. The primary difference between these LLM-DT methods and our LBM methods is that LLM-DTs do not incorporate the reasoning capabilities of LLMs.

For a detailed implementation of our LBM method's prompt design, we present a case in Fig. \ref{fig:inference}. At each decision timestep $t$
, the prompt is generated at timestep $t-1$. This prompt includes key history 4 timesteps' performance metrics such as achieved conversions, remaining budget, and predicted impression value (calculated as the average value of single impression opportunities multiplied by the number of impression opportunities). The LLM is then tasked with determining the CPA ratio using this information, which enhances the LBM-Think's understanding of the prompt in a self-supervised manner and can be used to assess any hallucinations. Finally, the LLM must reason about the adjustment direction for future bidding parameters.
Finally, at timestep $t$, the LBM-Act model receives this CoT and the previous 10 timesteps' numerical sequence information as inputs to generate the final action for execution, as illustrated in Fig. \ref{fig:inference}.

For detailed information on training the Q-value in the GQPO method, we utilize a transformer that receives a sequence of states and actions as inputs to build the Q-value. The hyperparameter details necessary for reproduction are listed in Table \ref{gas_hyperparameters}. The training method follows the IQL approach \cite{iql}.

\label{hyper_setting}
\begin{table}[h!]
\caption{The detailed hyperparameters of Q-value networks}
\scalebox{0.9}{
\begin{tabular}{c|c}
\hline
Hyperparameters     & Value  \\ \hline
Batch size          & 128    \\
Number of steps     & 400000 \\
Sequence length     & 10     \\
Learning rate       & 1e-4   \\
Number of attention layers       & 6   \\
Number of heads       & 8   \\
Optimizer           & AdamW  \\
Optimizer eps       & 1e-8   \\
Weight decay        & 1e-2   \\
Scale               & 2000   \\
Episode length      & 48     \\
Hidden size         & 512    \\
Activation function & ReLU   \\
Gamma               & 0.99   \\
Tau                 & 0.01   \\
Expectile           & 0.7    \\ \hline
\end{tabular}}
\label{gas_hyperparameters}
\end{table}

% \section{Evaluation on GRPO}
% We employ the GRPO method to fine-tune the LLM for generating CoT and actions, aiming to enhance its reasoning capabilities while keeping action generation ability in a single LLM.  Nonetheless, as illustrated in Table \ref{tab:gen_baseline}, this GRPO approach underperforms in auto-bidding, failing to fully utilize the budget. Furthermore, we noticed instability in GRPO's training for this task, frequently leading to shorter or missing CoT, as depicted in Fig. \ref{fig:train_response_length}.

% \begin{figure}[htbp]
%     \centering
%     \includegraphics[width=0.7\textwidth]{figs/train_response_length.pdf}
%     \caption{Finetune the LLM using GRPO so that it first performs step-by-step reasoning and then outputs an action. However, after around 150 training steps, the model only outputs the action without performing additional reasoning.}
%     \label{fig:train_response_length}
% \end{figure}

\subsection{Discussion on Online Learning and RL}
Online Learning (OL) is also an important learning method for addressing auto-bidding problems \cite{OL_R1, OL_R2, OL_R3}. OL primarily focuses on minimizing regret by typically selecting actions from a discrete feasible set while adhering to budget and ROI constraints for every timestep \cite{OL_R2}. In contrast, RL aims to maximize cumulative rewards by satisfying constraints at the end of the period and can operate in a continuous action space, typically implemented via neural networks, which offers potentially superior solutions. Additionally, OL estimates the optimal dual variables or bids under assumptions like stable and predictable cost and conversion distribution, while RL is free from these optimality assumptions by adjusting the bidding parameters, making it more adaptable to dynamic and complex environments. Moreover, OL typically requires policy refinement through exploration in the environment, which can cause risk concerns, similar to the challenges faced in online RL.
In contrast, offline RL directly learns policies from datasets, with state-of-the-art techniques including generative methods like Decision Transformer, Diffusers, and our LBM.
For the relationship between LBM and offline RL, intuitively, the core technique of LBM, GQPO, serves as an effective adaptation of AWR, transitioning from common numerical action spaces in offline RL tasks to language spaces, where GQPO treats CoT as the "action".

\end{document}